\documentclass[sigconf]{aamas}

\usepackage{balance} 
\usepackage{amsmath} 
\usepackage{pifont}
\usepackage{multirow}

\doi{LJRK3716}

\makeatletter
\gdef\@copyrightpermission{
  \begin{minipage}{0.2\columnwidth}
   \href{https://creativecommons.org/licenses/by/4.0/}{\includegraphics[width=0.90\textwidth]{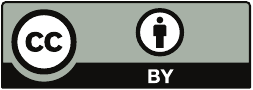}}
  \end{minipage}\hfill
  \begin{minipage}{0.8\columnwidth}
   \href{https://creativecommons.org/licenses/by/4.0/}{This work is licensed under a Creative Commons Attribution International 4.0 License.}
  \end{minipage}
  \vspace{5pt}
}
\makeatother

\setcopyright{ifaamas}
\acmConference[AAMAS '26]{Proc.\@ of the 25th International Conference
on Autonomous Agents and Multiagent Systems (AAMAS 2026)}{May 25 -- 29, 2026}
{Paphos, Cyprus}{C.~Amato, L.~Dennis, V.~Mascardi, J.~Thangarajah (eds.)}
\copyrightyear{2026}
\acmYear{2026}
\acmDOI{}
\acmPrice{}
\acmISBN{}

\title[AAMAS-2026 Formatting Instructions]{QDepth-VLA: Quantized Depth Prediction as Auxiliary Supervision for Vision-Language-Action Models}

\author{Yixuan Li}
\affiliation{
  \institution{School of Artificial Intelligence, University of Chinese Academy of Sciences}
  \city{Beijing}
  \country{China}}
\email{liyixuan223@mails.ucas.ac.cn}

\author{Yuhui Chen}
\affiliation{
  \institution{Institute of Automation, Chinese Academy of Sciences}
  \city{Beijing}
  \country{China}}
\affiliation{
  \institution{School of Artificial Intelligence, University of Chinese Academy of Sciences}
  \city{Beijing}
  \country{China}}
\email{chenyuhui2022@ia.ac.cn}

\author{Mingcai Zhou}
\affiliation{
  \institution{Institute of Automation, Chinese Academy of Sciences}
  \city{Beijing}
  \country{China}}
\affiliation{
  \institution{Beijing Zhongke Huiling Robot Technology Co.}
  \city{Beijing}
  \country{China}}
\email{mingcai.zhou@ia.ac.cn}

\author{Zhengtao Zhang}
\affiliation{
  \institution{Institute of Automation, Chinese Academy of Sciences}
  \city{Beijing}
  \country{China}}
\affiliation{
  \institution{Beijing Zhongke Huiling Robot Technology Co.}
  \city{Beijing}
  \country{China}}
\email{zhengtao.zhang@ia.ac.cn}

\author{Dongbin Zhao}
\affiliation{
  \institution{Institute of Automation, Chinese Academy of Sciences}
  \city{Beijing}
  \country{China}}
\affiliation{
  \institution{School of Artificial Intelligence, University of Chinese Academy of Sciences}
  \city{Beijing}
  \country{China}}
\email{dongbin.zhao@ia.ac.cn}

\author{Haoran Li}\authornote{Corresponding author.}
\affiliation{
  \institution{Institute of Automation, Chinese Academy of Sciences}
  \city{Beijing}
  \country{China}}
\affiliation{
  \institution{School of Artificial Intelligence, University of Chinese Academy of Sciences}
  \city{Beijing}
  \country{China}}
\email{lihaoran2015@ia.ac.cn}

\begin{abstract}
Spatial perception and reasoning are crucial for Vision–Language–
Action (VLA) models to accomplish fine-grained manipulation tasks. However, existing approaches often lack the ability to understand and reason over the essential 3D structures necessary for precise control. To address this limitation, we propose QDepth-VLA, a general framework that augments VLA models with an auxiliary depth prediction task. A dedicated depth expert is designed to predict quantized latent tokens of depth maps obtained from a VQ-VAE encoder, enabling the model to learn depth-aware representations that capture critical geometric cues. Experimental results on the simulation benchmarks and real-world tasks demonstrate that QDepth-VLA yields strong spatial reasoning and competitive performance on manipulation tasks.
\end{abstract}

\keywords{Vision–Language–Action models; Quantized depth prediction; Spatial reasoning; Robotic manipulation}

\newcommand{\BibTeX}{\rm B\kern-.05em{\sc i\kern-.025em b}\kern-.08em\TeX}
\newcommand{\xmark}{\ding{55}}

\begin{document}


\pagestyle{fancy}
\fancyhead{}


\maketitle 


\section{Introduction}

Large Vision-Language-Action (VLA) models \cite{Black20240AV, Intelligence202505AV, Cheang2024GR2AG, Cheang2025GR3TR, RoboGPT} have recently emerged as a powerful paradigm for robotic learning. By grounding pre-trained Vision-Language Models (VLMs)  \cite{Beyer2024PaliGemmaAV, Steiner2024PaliGemma2A, Wang2024Qwen2VLEV} with action-generation capabilities, robots acquire strong generalization across diverse instructions and visual contexts \cite{ChenY1-RSS-25}. However, when applied to long-horizon or fine-grained manipulation tasks, these models often exhibit substantial performance degradation \cite{Yang2025RoboEnvisionAL, Fan2025LongVLAUL, Song2025ReconVLARV, Zhu2025BridgingVA}.
The primary reason lies in a persistent gap between semantic understanding and geometric reasoning \cite{Goyal2024RVT2LP, Gervet2023Act3D3F}.

Without reliable 3D understanding, VLAs often misestimate object positions or gripper-object relations, leading to cascading errors during manipulation \cite{Song2025ReconVLARV}.
Therefore, several recent works have explored incorporating geometric information into VLA models to enable a deeper understanding of the 3D physical environment. 
These approaches can be broadly generalized into three paradigms: direct 3D feature injection, 2D-projected 3D feature integration and auxiliary 3D information prediction. 
The first category injects encoded 3D representations, such as point clouds \cite{Li2025PointVLAIT} or depth maps \cite{Bhat20253DCL}, into the vision–language backbone or the action head. 
This strategy typically requires an additional encoder to process 3D features, increasing model complexity and computational cost. 
While providing explicit geometric cues, it may disrupt the powerful 2D priors learned during large-scale VLM pretraining, leading to degraded visual-language reasoning and understanding. 
The second category projects 3D features into 2D representations and feeds them into the VLM \cite{Li2025BridgeVLAIA}. 
Although this preserves pretrained 2D priors, it inevitably introduces information loss in the projection process, which can hinder fine-grained manipulation performance. 
Compared to these two paradigms, enhancing geometric understanding through auxiliary visual prediction tasks, such as future depth maps estimation  \cite{Zhang2025DreamVLAAV}, offers a more promising alternative. 
This approach not only preserves the strong 2D priors of pretrained VLMs, but also avoids the need for additional sensory inputs during inference, while encouraging the model to learn 3D-consistent spatial reasoning. 

However, existing works that employ depth-map-based visual prediction as auxiliary tasks \cite{Zhang2025DreamVLAAV} have not achieved consistent performance improvements, and in some cases even indicate that introducing depth prediction as an auxiliary loss can be detrimental to policy learning due to noisy supervision and weak geometric grounding. 
The key challenges lie in three aspects.
Firstly, the supervision quality of depth maps is often limited by insufficient spatial–temporal consistency across frames \cite{Yang2024DepthAV, Chen2025VideoDA}, introducing substantial noise that weakens geometric grounding. 
Secondly, pixel-wise depth regression produces highly redundant learning signals, forcing the model to reconstruct every pixel rather than focusing on salient structural cues essential for manipulation. 
Thirdly, using a vision-language backbone to predict depth maps may interfere with its pre-trained semantic alignment, potentially degrading multimodal reasoning performance.

To address these challenges, we propose \textbf{QDepth-VLA}\footnote{Supplementary material and source code are publicly available at:
\url{https://github.com/ucasmichael/QDepth-VLA}.}, which augments large VLAs by introducing quantized depth prediction as an auxiliary supervision signal.
Instead of regressing pixel-wise depth values, QDepth-VLA learns discrete depth representations through vector quantization, capturing salient structural information in a compact and optimization-friendly manner. 
An independent depth expert is also introduced to predict these quantized depth tokens, enabling the model to leverage geometric cues without interfering with the vision-language backbone’s pretrained semantic alignment.
Our main contributions are summarized as follows:

\begin{enumerate}
\item We introduce QDepth-VLA, a novel VLA model enhanced with quantized depth information. By integrating a depth prediction task, it internalizes geometric understanding, enabling more accurate reasoning about object spatial relationships.
\item To facilitate more robust depth learning, we design a specialized \emph{Depth Expert} that predicts quantized depth tokens rather than raw pixel-level depth maps. This formulation effectively mitigates the impact of depth noise and provides a more compact, optimization-friendly supervision signal for geometry-aware policy learning.
\item Comprehensive experiments on both the Simpler \cite{Li2024EvaluatingRR} and LIBERO \cite{Liu2023LIBEROBK} benchmarks demonstrate that QDepth-VLA substantially enhances policy performance, outperforming Open $\pi_0$ \cite{Ren2024OpenPiZero} by 6.1\% and 7.7\% on average success rate, respectively. Moreover, QDepth-VLA achieves a 10.0\% improvement in real-world robotic manipulation, validating its effectiveness and generalizability.
\end{enumerate}


\section{RELATED WORKS}

\subsection{3D-Enhanced VLA}

3D spatial information has been widely explored to overcome the limitations of purely 2D-based models. Early efforts typically enhanced spatial perception by either lifting 2D inputs into 3D  \cite{Gervet2023Act3D3F, Goyal2023RVTRV, Goyal2024RVT2LP} or directly fusing 2D visual features with 3D point clouds \cite{Yang2025FP3A3, Qu2025SpatialVLAES, Lin2025Evo0VM}.

While these approaches demonstrate that incorporating 3D signals can significantly improve spatial perception and action precision, directly fusing 3D and 2D representations or relying solely on 3D features can disrupt the visual-language alignment established in large-scale VLM pretraining. To mitigate this, two alternative directions have been proposed:
(1) Projecting 3D features into 2D space, as in BridgeVLA  \cite{Li2025BridgeVLAIA}, which renders 3D inputs into multi-view 2D images for compatibility with VLMs.
(2) Independent 3D encoders encode geometric information for integration into the action head. This paradigm is employed by PointVLA \cite{Li2025PointVLAIT} and GeoVLA \cite{Sun2025GeoVLAE3}, where specialized point cloud encoders supply 3D embeddings to modality-specific experts.

Despite these advances, point cloud reconstruction may lose fine-grained object details, and the modality gap between 2D RGB pretraining and 3D geometry remains a persistent challenge. By contrast, depth maps exhibit a much smaller gap with RGB images and thus offer a more natural bridge between 2D and 3D. Recent depth-based approaches have demonstrated this advantage. 3D-CAVLA \cite{Bhat20253DCL} integrates Region of Interest (RoI) pooling with depth embeddings projected into VLM token space, achieving extraordinary multi-view performance, while 4D-VLA \cite{Zhang20254DVLASV}  augments visual inputs with 3D coordinate embeddings to support both spatial alignment and temporal reasoning.

Motivated by these insights, we adopt depth maps as the 3D augmentation source. Crucially, instead of directly fusing them with RGB features which risks interfering with pre-trained VLM semantics, we reformulate depth as an auxiliary prediction task. This design enables QDepth-VLA to move beyond passive depth perception toward depth understanding, a capability we elaborate on in the next section.

\subsection{Auxiliary Visual Reasoning Tasks for VLA}
While depth maps offer a natural bridge between 2D and 3D for enhancing spatial grounding, another promising direction is to strengthen the reasoning capacity of VLAs through auxiliary visual prediction tasks. Instead of passively mapping inputs to actions, policies can be trained to output intermediate signals that make future-oriented reasoning explicit, thereby providing richer supervision during training and improving long-horizon planning at inference.

A series of works focus on predicting future sub-goals, such as generating sub-goal images or short rollouts that visualize task progress. This strategy, as exemplified by CoT-VLA \cite{Zhao2025CoTVLAVC} , enhances temporal reasoning by conditioning action generation on both current and predicted states, but incurs high computational cost due to the difficulty of synthesizing realistic RGB predictions. Other researches \cite{Intelligence202505AV, Cheang2025GR3TR} introduce object-centric signals, such as bounding boxes or spatial relations, which provide structured knowledge of entities and their interactions. More recently, latent future embeddings have been explored, where discrete action tokens predicted in a compressed latent space encode upcoming intentions. AgiBot World Colosseo \cite{AgiBotWorldContributors2025AgiBotWC} and UniVLA \cite{Bu2025UniVLALT} exemplify this paradigm, showing scalability through large-scale human video pretraining, yet such latent predictions often lack explicit 3D grounding and struggle to capture fine-grained geometry. Finally, some approaches turn to pixel-level 3D supervision, predicting dense depth or semantic maps to reinforce geometric awareness, as in 3D-VLA \cite{Zhen20243DVLAA3} and DreamVLA \cite{Zhang2025DreamVLAAV}. While sometimes effective for strengthening spatial reasoning, these signals are difficult to optimize directly and may overemphasize redundant low-level cues rather than the most relevant spatial structures.

Different from previous works, our approach unifies 3D information enhancement and visual reasoning by introducing depth codebook prediction—an auxiliary task that brings 3D cues into reasoning in a compact and semantically meaningful way, while remaining naturally aligned with language-conditioned action policies.


\section{Methodology}

\begin{figure*}[thpb]
    \centering
    \includegraphics[width=\textwidth]{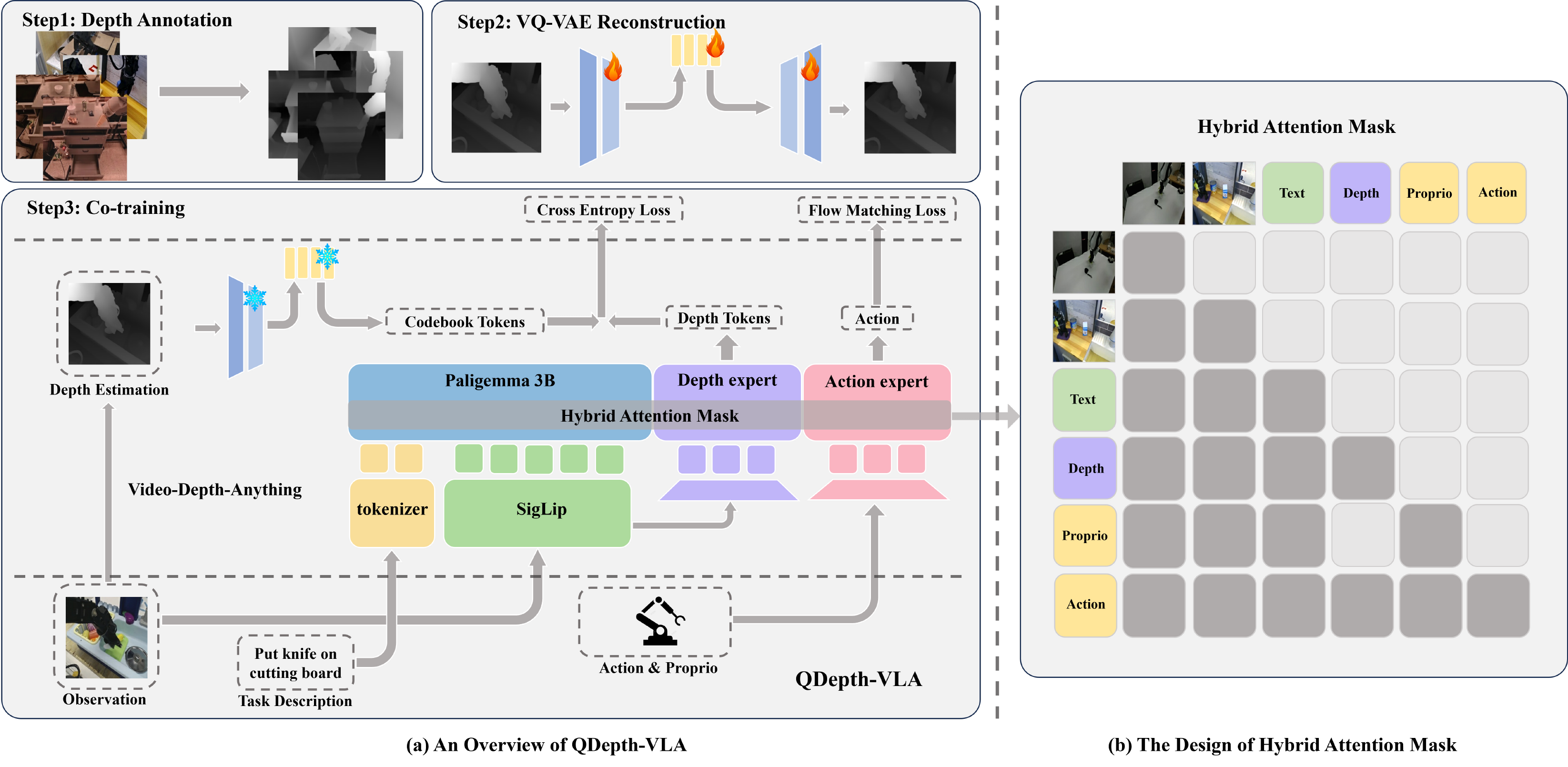}
    \caption{An overview of QDepth-VLA. (a) The overall architecture and training pipeline, where depth supervision is incorporated via a depth expert and latent prediction module. In co-training, the VQ-VAE  \cite{Oord2017NeuralDR} encoder and codebook are frozen, while PaLI-Gemma 3B \cite{Beyer2024PaliGemmaAV}, the action expert, depth expert, SigLIP \cite{Zhai2023SigmoidLF}, and tokenizer are trainable. (b) The proposed hybrid attention mask, which integrates depth and visual tokens to enhance spatial reasoning and manipulation performance.}
    \label{fig:model_arch}
    \Description{Overview diagram of the QDepth-VLA model. Panel (a) illustrates the full architecture and training pipeline. Visual inputs are encoded into visual tokens and processed by a vision-language backbone. A depth expert predicts quantized depth tokens using a latent prediction module, while an action expert predicts robot actions. The VQ-VAE encoder and codebook are frozen during co-training, and other modules are trainable. Arrows indicate the data flow between components. Panel (b) shows a hybrid attention mask that combines visual tokens and depth tokens, enabling interaction between geometric and visual information for improved spatial reasoning and manipulation.}
\end{figure*}

\subsection{Depth Annotation}
Since existing VLA datasets such as OXE dataset \cite{Padalkar2023OpenXR}  lack sufficient 3D annotations, we first generate monocular depth estimates for training.
To ensure high-quality and spatial–temporal consistent depth sequences, we employ \textit{Video-Depth-Anything} (ViDA) \cite{Chen2025VideoDA}, the current state-of-the-art monocular video depth estimation framework built upon a ViT-Large backbone, to acquire depth maps.
Specifically, ViDA is applied to the main-view RGB frames from a subset of the OXE \cite{Padalkar2023OpenXR} and LIBERO \cite{Liu2023LIBEROBK}  datasets to obtain temporally aligned relative depth annotations, providing reliable geometric supervision for depth tokenization and subsequent model training.

\subsection{VQ-VAE Reconstruction}
To represent depth compactly, we pretrain a Vector-Quantized Variational Autoencoder (VQ-VAE) \cite{Oord2017NeuralDR}.  
Given a depth frame $\mathbf{x}$, the encoder $f_\theta(\cdot)$ produces a latent $\mathbf{z}_e = f_\theta(\mathbf{x})$, which is quantized to the nearest code vector in a codebook $\mathcal{C} = \{\mathbf{c}_1, \dots, \mathbf{c}_K\}$:
\begin{eqnarray}
    \mathbf{z}_q = \mathbf{c}_{j^*}, \quad 
    j^* = \arg\min_j \|\mathbf{z}_e - \mathbf{c}_j\|_2^2.
\end{eqnarray}
We use $K = 256$ codebook entries of dimension $d = 160$, and train the VQ-VAE \cite{Oord2017NeuralDR} with the standard objective:
\begin{eqnarray}
    \mathcal{L}_{\text{vq}} = 
    \underbrace{\ell_{\text{rec}}(\mathbf{x}, g_\phi(\mathbf{z}_q))}_{\text{reconstruction}} 
    + \underbrace{\|\text{sg}[\mathbf{z}_e] - \mathbf{c}_{j^*}\|_2^2}_{\text{codebook update}} 
    + \beta \underbrace{\|\mathbf{z}_e - \text{sg}[\mathbf{c}_{j^*}]\|_2^2}_{\text{commitment}},
\end{eqnarray}
where $\text{sg}[\cdot]$ denotes stop-gradient and $\beta = 0.25$.  
In practice, we experiment with latent grid resolutions of $16 \times 16$ and $32 \times 32$.
We find that the smaller $16 \times 16$ configuration already achieves accurate depth reconstruction while remaining computationally efficient.
The VQ-VAE \cite{Oord2017NeuralDR} is pretrained independently on each dataset using AdamW \cite{Loshchilov2017DecoupledWD} with a learning rate of $1\times10^{-5}$ to ensure stable convergence and reconstruction quality.
The resulting pretrained model produces discretized depth code indices, which serve as supervisory targets for the depth expert in QDepth-VLA.

\subsection{QDepth-VLA Architecture}

\begin{table}[t]
\centering
\caption{Key configurations of the Action and Depth experts of QDepth-VLA.}
\label{tab:expert_configs}
\begin{tabular}{|l|cc|}
\toprule
 & \multicolumn{2}{c|}{\textbf{QDepth-VLA}}  \\
\cmidrule(lr){2-3}
 & \textbf{Action Expert} & \textbf{Depth Expert} \\
\midrule
Backbone       & Transformer      & Transformer     \\
Layers / Heads & 18 / 8           & 18 / 8          \\
Hidden dim     & 1024             & 1024           \\
Interm. dim    & 4096             & 4096            \\
Inputs         & Proprio + Action & RGB-Img tokens    \\
Outputs        & Actions           & Depth tokens     \\
\bottomrule
\end{tabular}
\end{table}

QDepth-VLA adopts a unified and modular architecture built upon Open $\pi_0$ \cite{Ren2024OpenPiZero}, extending its VLA pipeline with an additional depth supervision branch.
As shown in Figure ~\ref{fig:model_arch}(a), the model consists of three parameterized modules: a pretrained vision–language model (VLM), an action expert, and a newly introduced \emph{depth expert}.
These modules are coordinated through a mixture-of-experts (MoE) structure and a carefully designed hybrid attention mask, enabling QDepth-VLA to jointly reason about geometry and control variants without disrupting pretrained representations.

We choose PaliGemma-3B \cite{Beyer2024PaliGemmaAV} as VLM backbone, which integrates SigLIP-based \cite{Zhai2023SigmoidLF} vision encoding with Gemma’s \cite{Mesnard2024GemmaOM} language modeling capability.
Input instructions are first tokenized using Gemma’s \cite{Mesnard2024GemmaOM} tokenizer, while the main-view RGB image is processed by the SigLIP \cite{Zhai2023SigmoidLF} image encoder to obtain 256 visual tokens.
These image tokens are concatenated with 20 text prefix tokens and fed into the Gemma \cite{Mesnard2024GemmaOM} decoder under full block attention to produce multimodal embeddings that capture both spatial and semantic cues.
This pretrained VLM remains trainable during the training stage, allowing geometric adaptation to manipulation environment.

The action expert is a transformer-based module responsible for translating multimodal embeddings and proprioceptive states into executable robot actions.
It consists of stacked transformer layers with MLP-based encoders and decoders that integrate visual–language context from the VLM with proprioceptive features.
This module, which is built upon the original Open $\pi_0$ \cite{Ren2024OpenPiZero} action head, functions as the core control head of QDepth-VLA.

To incorporate geometric reasoning, QDepth-VLA introduces a dedicated depth expert, architecturally aligned with the action expert (illustrated in Table~\ref{tab:expert_configs}).
It takes the visual embeddings from the SigLIP \cite{Zhai2023SigmoidLF} encoder as input, before language fusion to avoid semantic interference.
These embeddings are projected through a lightweight MLP, processed by a transformer backbone, and then passed to a shallow CNN decoder that predicts 256 depth tokens. 
Each predicted token corresponding to a latent vector is then aligned with the quantized tokens produced by the pretrained VQ-VAE \cite{Oord2017NeuralDR} encoder over its codebook.
The pretrained VQ-VAE \cite{Oord2017NeuralDR} decoder is subsequently used to reconstruct the spatial depth map from these latent tokens when required.
This discrete formulation enables QDepth-VLA to capture compact, structured geometric representations while maintaining optimization stability.

As for hybrid attention mechanism, existing designs typically employ a standard causal attention structure, as seen in DreamVLA \cite{Zhang2025DreamVLAAV} and CoT-VLA \cite{Zhao2025CoTVLAVC}. However, since depth modalities inherently contain noise, directly fusing them under causal attention may introduce undesirable interference, potentially degrading action generation quality \cite{Zhang2025DreamVLAAV}.
To address this issue, we redesign the hybrid attention mechanism (Figure ~\ref{fig:model_arch}(b)) to more effectively regulate cross-modal information flow among text, image, depth, proprioception, and action tokens. To be specific:

(1) Text and image tokens attend only within their modality to preserve pretrained semantic grounding.

(2) Depth tokens attend to both image and text tokens, contextualizing geometric features with visual semantics.

(3) Action tokens attend to all preceding modalities, integrating fused perceptual and geometric cues for policy generation.

This hierarchical attention design allows depth to enhance spatial understanding while preventing over-interference with the pretrained VLM and keeping computation efficient.

\subsection{Co-Training Procedures}

\subsubsection{Quantized Depth Supervision}

During joint training, the depth expert predicts latent depth tokens, and these tokens are then used to compute logits with the encoded image features over the VQ-VAE \cite{Oord2017NeuralDR} codebook:
\begin{eqnarray}
\mathscr{l}_{i,k} = -\frac{1}{\tau}\left\|x_i - c_k\right\|_2^2,
\end{eqnarray}
where $i$ indexes latent spatial positions , $k$ indexes codebook entries ($K=256$) and $\tau$ represents temperature factor.  
A cross-entropy loss is applied using ground-truth code indices $z_i^*$ obtained from the pretrained VQ-VAE \cite{Oord2017NeuralDR}:
\begin{eqnarray}
\mathcal{L}_{\text{depth}} = 
-\frac{1}{B \cdot N}\sum_{i=1}^{B \cdot N}
\log\frac{\exp(\mathscr{l}_{i, z_i^*})}{\sum_{k=1}^{K}\exp(\mathscr{l}_{i,k})},
\end{eqnarray}
where $B$ is batch size and $N$ the number of latent tokens per frame.  
This loss encourages the visual encoder to learn geometry-aware embeddings aligned with the quantized depth representation.

\subsubsection{Action Modeling}

Based on the underlying VLA backbone, the action prediction objective is as follows:

The Conditional Flow Matching (CFM) action loss \cite{Lipman2022FlowMF}  is identical to that of $\pi_0$ \cite{Black20240AV}:
    \begin{eqnarray}
    \mathcal{L}_{\text{CFM}}(\theta)
    = 
    \mathbb{E}_{p(A_t|O_t),\, q(\hat{A}_t^{\lambda}|A_t)}
    \| f_\theta(\hat{A}_t^{\lambda}, O_t) - g(\hat{A}_t^{\lambda}|A_t) \|_2^2 ,
    \end{eqnarray}
    where the action chunk
$A_t = [a_t, a_{t+1}, \ldots, a_{t+H-1}]$
is conditioned on the observation
$O_t = [I_t, \ell_t, s_t]$,
which includes the RGB image, language instruction, and end-effector state.
Notably, $\hat{A}_t^{\lambda}$ denotes noisy action samples generated from a diffusion-like process:
\begin{eqnarray}
\hat{A}_t^{\lambda} = \lambda A_t + (1-\lambda)\eta , \quad
\eta \sim \mathcal{N}(0,I),
\end{eqnarray}
and the corresponding noise distribution and flow target are defined as:
\begin{eqnarray}
q(\hat{A}_t^{\lambda} \mid A_t) = \mathcal{N}(\lambda A_t, (1-\lambda)I), \quad
g(\hat{A}_t^{\lambda} \mid A_t) = \eta - A_t.
\end{eqnarray}
This formulation enables the model to approximate a continuous-time flow field that transports noisy actions toward their clean ground-truth counterparts.

\subsubsection{Co-Training Objectives}

The total loss combines the action and depth objectives:
\begin{eqnarray}
\mathcal{L}_{\text{total}} = 
\mathcal{L}_{\text{action}} + \lambda_t \cdot \mathcal{L}_{\text{depth}},
\end{eqnarray}
where $\lambda_t = \lambda_0 \cdot \gamma^t$ exponentially decays over training steps, with $\lambda_0=0.01$.  
This co-training schedule enables the model to first establish stable geometric alignment before gradually focusing on action refinement.

\begin{table*}[t]
\centering
\caption{Results of QDepth-VLA on the LIBERO benchmark.}
\label{tab:libero_results_combined}
\begin{tabular}{l|l|l|cccc|c}
\toprule
\textbf{View Setting} & \textbf{Category} & \textbf{Method}  & \textbf{Spatial} & \textbf{Object} & \textbf{Goal} & \textbf{Long} & \textbf{Avg} \\
\midrule
\multirow{6}{*}{\textbf{Single-view VLA}} 
 & \multirow{3}{*}{General VLA} 
 & OpenVLA finetuned \cite{Kim2024OpenVLAAO}  & 84.7 & 88.4 & 79.2 & 53.7 & 76.5 \\
 &  & CoT-VLA-7B \cite{Zhao2025CoTVLAVC}  & 87.5 & 91.6 & 87.6 & 69.0 & 81.1 \\
 &  & Open $\pi_0$ \cite{Ren2024OpenPiZero}  & 77.2 & 84.0 & 83.6 & 66.0 & 77.7 \\
 \cmidrule(lr){2-8}
 & 3D-cloud-enhanced VLA 
 & SpatialVLA \cite{Qu2025SpatialVLAES}   & \textbf{88.2} & 89.9 & 78.6 & 55.5 & 78.1 \\
 \cmidrule(lr){2-8}
 & \multirow{2}{*}{Depth-enhanced VLA} 
 & 3D-CAVLA \cite{Bhat20253DCL}  & 86.1 & \textbf{94.7} & 82.9 & 66.8 & 82.6 \\
 &  & \textbf{QDepth-VLA (ours)}  & 86.0 & 88.8 & \textbf{94.0} & \textbf{72.6} & \textbf{85.4} \\
\midrule
\multirow{10}{*}{\textbf{Multi-view VLA}} 
 & \multirow{5}{*}{General VLA} 
 & Diffusion Policy \cite{Chi2023DiffusionPV}  & 78.3 & 92.5 & 68.3 & 50.5 & 72.4 \\
 &  & Octo finetuned \cite{Team2024OctoAO}  & 78.9 & 85.7 & 84.6 & 51.1 & 75.1 \\
 &  & $\pi_0$-FAST finetuned \cite{Black20240AV}  & 96.4 & 96.8 & 88.6 & 60.2 & 85.5 \\
 &  & $\pi_0$ finetuned \cite{Black20240AV}  & 96.8 & 98.8 & 95.8 & 85.2 & 94.2 \\
 &  & UniVLA \cite{Bu2025UniVLALT}  & 96.5 & 96.8 & 95.6 & 92.0 & 95.2 \\
 \cmidrule(lr){2-8}
 & 3D-cloud-enhanced VLA 
 & GeoVLA \cite{Sun2025GeoVLAE3}  & \textbf{98.4} & 99.0 & 96.6 & \textbf{96.6} & 97.7 \\
 \cmidrule(lr){2-8}
 & \multirow{4}{*}{Depth-enhanced VLA} 
 & 3D-CAVLA \cite{Bhat20253DCL}  & 98.2 & \textbf{99.8} & \textbf{98.2} & 96.1 & \textbf{98.1} \\
 &  & 4D-VLA \cite{Zhang20254DVLASV}  & 88.9 & 95.2 & 90.9 & 79.1 & 88.6 \\
 &  & DreamVLA \cite{Zhang2025DreamVLAAV}  & 97.5 & 94.0 & 89.5 & 89.5 & 92.6 \\
 &  & \textbf{QDepth-VLA (ours)} & 97.6 & 96.6 & 95.2 & 90.0 & 94.9 \\
\bottomrule
\end{tabular}
\end{table*}

\subsubsection{Optimization Setup}

QDepth-VLA is trained using the AdamW \cite{Loshchilov2017DecoupledWD}  optimizer with decoupled weight decay. 
We set the learning rate for both the action expert and the VLM backbone to $5 \times 10^{-5}$. 
A cosine learning rate scheduler with $200$ warm-up steps and a cycle length of $10^7$ steps is applied, ensuring stable optimization throughout training.


\section{EXPERIMENTS}

In this section, we conduct comprehensive experiments across both simulation and real-world settings to evaluate the effectiveness of our approach. Specifically, we aim to address the following three questions: 

(1) Can depth supervision effectively enhance VLA performance in long-horizon and pick-and-place tasks, particularly those requiring fine-grained manipulation?  

(2) Is depth supervision more effective than pixel-level depth prediction?  

(3) Does the proposed hybrid attention mask contribute to performance gains?

\subsection{Simulation Experiments}

\subsubsection{Training Recipe}

The QDepth-VLA based on Open $\pi_0$ \cite{Ren2024OpenPiZero} is initially pre-trained for 9 epochs on the Fractal dataset \cite{Brohan2022RT1RT}, followed by 20 epochs of pre-training on the LIBERO-90 dataset  \cite{Liu2023LIBEROBK}.
After pre-training, the model is further fine-tuned on the four LIBERO subsets — Spatial, Object, Goal and Long  \cite{Liu2023LIBEROBK} for around 50 epochs.
For the Simpler benchmark \cite{Li2024EvaluatingRR}, the model is instead trained from scratch, first using the Bridge dataset  \cite{Walke2023BridgeDataVA} for 13 epochs, and then the Fractal dataset  \cite{Brohan2022RT1RT} for an additional 9 epochs.

All experiments are conducted using the Fully Sharded Data Parallel (FSDP) training strategy on 8 × NVIDIA H20 GPUs.
A per-GPU batch size of 32 is used, yielding a global batch size of 1024 with gradient accumulation, and the action chunk size is fixed at 4.

\subsubsection{Evaluation Setup}
For evaluation on LIBERO \cite{Liu2023LIBEROBK}, we adopt its four benchmark suites (Spatial, Object, Goal and Long). Following the preprocessing method in \cite{Kim2024OpenVLAAO}, image resolution is first normalized to $256 \times 256$ and then resized to $224 \times 224$ as model input. We also apply a 180-degree rotation to all images and use only the main-view RGB observations. Each task is evaluated over 50 rollouts, with the average success rate reported.

On the Simpler benchmark \cite{Li2024EvaluatingRR}, the evaluation covers two distinct settings: (1) models trained on the Bridge dataset  \cite{Walke2023BridgeDataVA} are tested on tasks involving the WidowX250 robot, and (2) models trained on the Fractal dataset  \cite{Brohan2022RT1RT} are tested on tasks for the Google Robot. We adopt the visual matching configuration from Simpler  \cite{Li2024EvaluatingRR}, evaluating each task across multiple initial positions with 10 rollouts per configuration. Depending on the number of configurations, the total number of evaluations per task ranges from 240 to 2400.

\subsubsection{Main Results}

\begin{table*}[t]
\centering
\caption{Results of QDepth-VLA on Simpler benchmark(Google Robot tasks)}
\label{tab:fractal_results}
\begin{tabular}{l|cccc|c}
\toprule
 & \textbf{Pick Coke Can} & \textbf{ Move Near} & \textbf{Open/Close Drawer}  & \textbf{Open Top Drawer and Put Apple In} & \textbf{Avg} \\
\midrule
RT-2-X \cite{Brohan2023RT2VM}       & 78.7 & 77.9  &  25.0 & - & 60.7 \\
Octo-Base \cite{Team2024OctoAO}    & 17.0 & 4.2 & 22.7 & - & 16.8 \\
OpenVLA \cite{Kim2024OpenVLAAO}      & 16.3 & 46.2 & 35.6 & - & 27.7 \\
RoboVLM finetuned \cite{Li2024TowardsGR} & 77.3 & 61.7 & 43.5 & - & 63.4 \\ 
SpatialVLA finetuned \cite{Qu2025SpatialVLAES} & 86.0 & 77.9 & 57.4 & - & 75.1 \\
Open $\pi_0$ \cite{Ren2024OpenPiZero} & 97.5 & \textbf{87.1} & \textbf{68.0} & 32.9 & 71.4 \\
\textbf{QDepth-VLA (ours)} & \textbf{98.3} & 81.4 & 58.0 & \textbf{62.6} &\textbf{75.1} \\
\bottomrule
\end{tabular}
\end{table*}

\begin{table*}[t]
\centering
\caption{Results of QDepth-VLA on Simpler benchmark(WidowX250 Robot tasks)}
\label{tab:bridge_results}
\begin{tabular}{l|cccc|c}
\toprule
 & \textbf{Put Carrot on Plate} & \textbf{Put Eggplant in  Basket} & \textbf{Put Spoon on Towel}  & \textbf{ Stack Block} & \textbf{Avg} \\
\midrule
Octo-Base \cite{Team2024OctoAO}    & 8.3 & 43.1 & 12.5 & 0.0 & 16.0 \\
OpenVLA \cite{Kim2024OpenVLAAO}      & 0.0 & 4.1 & 0.0 & 0.0 & 1.0 \\
RoboVLM finetuned \cite{Li2024TowardsGR} & 25.0 & 58.3 & 29.2 & 12.5 & 31.3\\ 
SpatialVLA finetuned \cite{Qu2025SpatialVLAES} & 25.0 & \textbf{100.0} & 16.7 & 29.2 & 42.7 \\
Open $\pi_0$ \cite{Ren2024OpenPiZero}  & \textbf{61.3}&89.6 &73.7 &15.8 &60.0 \\
\textbf{QDepth-VLA (ours)} & 57.5& 95.0 &\textbf{82.0}&\textbf{39.6}&\textbf{68.5} \\
\bottomrule
\end{tabular}
\end{table*}

\paragraph{\textbf{LIBERO Benchmark}}

QDepth-VLA adopts a \textit{single-view} setting, where the visual input consists of only one RGB image.
This contrasts with \textit{multi-view} models, which take multiple images as input, including temporally adjacent frames from historical observations.

As shown in Table~\ref{tab:libero_results_combined}, QDepth-VLA consistently outperforms single-view baselines across the LIBERO \cite{Liu2023LIBEROBK} suites.
It achieves stronger performance on both fine-grained and long-horizon tasks, reaching 94.0\% on the Goal tasks and 72.6\% on the Long tasks, surpassing the single-view baseline CoT-VLA  \cite{Zhao2025CoTVLAVC} by 6.4\% and 3.6\%, respectively.
Compared with Open $\pi_0$ \cite{Ren2024OpenPiZero}, QDepth-VLA shows consistent improvements across all four subsets (Spatial, Object, Goal and Long), with the largest gain of 8.8\% observed on the Spatial tasks.

While QDepth-VLA operates with only a single RGB observation, its average success rate remains competitive with multi-view VLAs.
Specifically, QDepth-VLA achieves a mean success rate only 0.1\% lower than $\pi_0$-FAST \cite{Ren2024OpenPiZero}, while exceeding 4D-VLA \cite{Zhang20254DVLASV} by 3.1\% and DreamVLA \cite{Zhang2025DreamVLAAV}  by 4.5\% on the Goal tasks.
Moreover, it surpasses $\pi_0$-FAST \cite{Black20240AV} by 12.4\% on the more challenging Long tasks.
Although leading multi-view models such as 3D-CAVLA \cite{Bhat20253DCL},  GeoVLA \cite{Sun2025GeoVLAE3} and UniVLA \cite{Bu2025UniVLALT} achieve higher overall results, our experimental results demonstrate that depth-augmented supervision effectively compensates for the lack of multi-view observations and brings single-view VLAs closer to multi-view performance levels.

By extension, we further implement a \textit{multi-view} variant of QDepth-VLA while maintaining the same setting that predicts latent depth tokens corresponding only to the current main-view image.
As shown in Table~\ref{tab:libero_results_combined}, the multi-view QDepth-VLA consistently outperforms single-view baselines.
It achieves an average success rate of 94.9\%, surpassing DreamVLA \cite{Zhang2025DreamVLAAV} by 0.1\% and $\pi_0$ \cite{Black20240AV} by 0.8\% on the Spatial tasks, and reaches 90.0\% success on the Long tasks.
While 3D-CAVLA \cite{Bhat20253DCL} and GeoVLA \cite{Sun2025GeoVLAE3} achieve higher average success rates, they require explicit point cloud or depth map inputs during inference — modalities that QDepth-VLA does not rely on.

These results reveal that the QDepth-VLA generalizes effectively to multi-view configurations, further enhancing geometric perception and long-horizon reasoning.

\paragraph{\textbf{Simpler Benchmark}}

Tables \ref{tab:fractal_results} and \ref{tab:bridge_results} present the experimental results of QDepth-VLA in the Simpler  \cite{Li2024EvaluatingRR} simulation environment.

As shown in Table~\ref{tab:fractal_results}, QDepth-VLA achieves a success rate of 98.3\% on the pick coke can task, surpassing Open $\pi_0$  \cite{Ren2024OpenPiZero} by 0.8\% and SpatialVLA  \cite{Qu2025SpatialVLAES} by 12.3\%. 
On the more complex open top drawer and put apple in task, QDepth-VLA  also attains 62.6\%, outperforming Open $\pi_0$ \cite{Ren2024OpenPiZero}  by a large margin of 29.7\%.
This substantial improvement on long-horizon tasks can be attributed to enhanced spatial perception and object localization provided by depth-guided supervision, which improves the model’s ability to accurately identify and grasp target objects.
As a result, the success probability of intermediate manipulation steps—such as grasping or placing—is increased, leading to higher overall task completion rates.

In Table~\ref{tab:bridge_results}, QDepth-VLA consistently achieves high success rates across various manipulation tasks.
Notably, on the stack block task—which demands precise spatial reasoning and fine-grained control—QDepth-VLA reaches a success rate of 39.6\%, surpassing SpatialVLA \cite{Qu2025SpatialVLAES} by 10.4\%.
Furthermore, on the Put Eggplant in Basket and Put Spoon on Towel tasks, QDepth-VLA achieves 95.0\% and 82.0\% success rates, respectively, outperforming Open $\pi_0$ \cite{Ren2024OpenPiZero} by 5.4\% and 8.3\%.
These improvements highlight the effectiveness of quantized depth supervision in enhancing spatial reasoning and manipulation precision, particularly for tasks involving object placement and coordination in cluttered 3D environments.

\paragraph{\textbf{Depth Reconstruction Visualization} }

\begin{figure*}[t]
    \centering
    \includegraphics[width=0.95\textwidth]{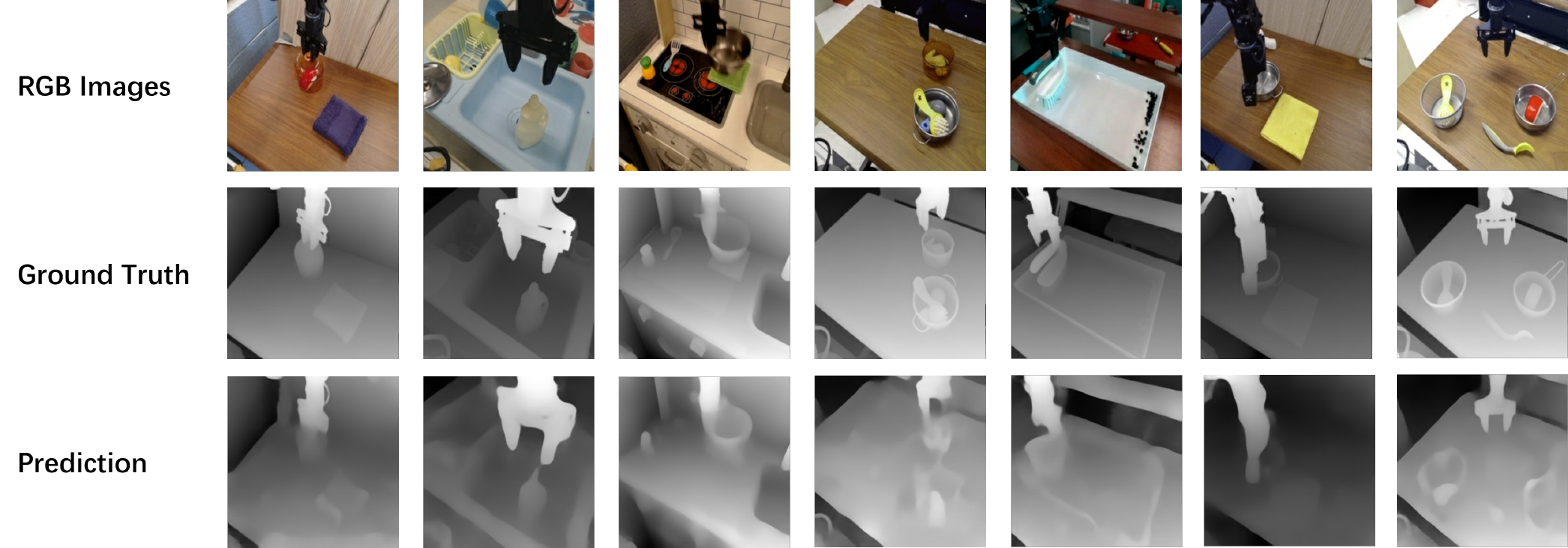}
    \caption{Depth reconstruction results from the QDepth-VLA. 
    Features predicted by depth expert are decoded using a trained VQ-VAE \cite{Oord2017NeuralDR} decoder. 
    The reconstructions demonstrate QDepth-VLA's ability to learn critical depth map features, including object and gripper boundaries, underscoring the success of its depth supervision.}
    \label{fig:depth_reconstruction}
    \Description{Visualization of depth reconstruction results. Each example shows an input image and the corresponding reconstructed depth map produced by decoding features predicted by the depth expert. Depth values are represented using a color scale, where different colors indicate different distances from the camera. The reconstructed maps highlight object shapes, table surfaces, and gripper boundaries, showing the preservation of geometric structure across different scenes.}
\end{figure*}

To further validate the effectiveness of our depth supervision, we visualize depth reconstructions by passing the quantized predicted features through a trained VQ-VAE \cite{Oord2017NeuralDR} decoder. As shown in Figure ~\ref{fig:depth_reconstruction}, the reconstructions preserve  structural details and align well with object boundaries, demonstrating that the learned depth representations capture spatial geometry in a meaningful way.

\begin{figure}[t]
    \centering
    \includegraphics[width=0.475\textwidth]{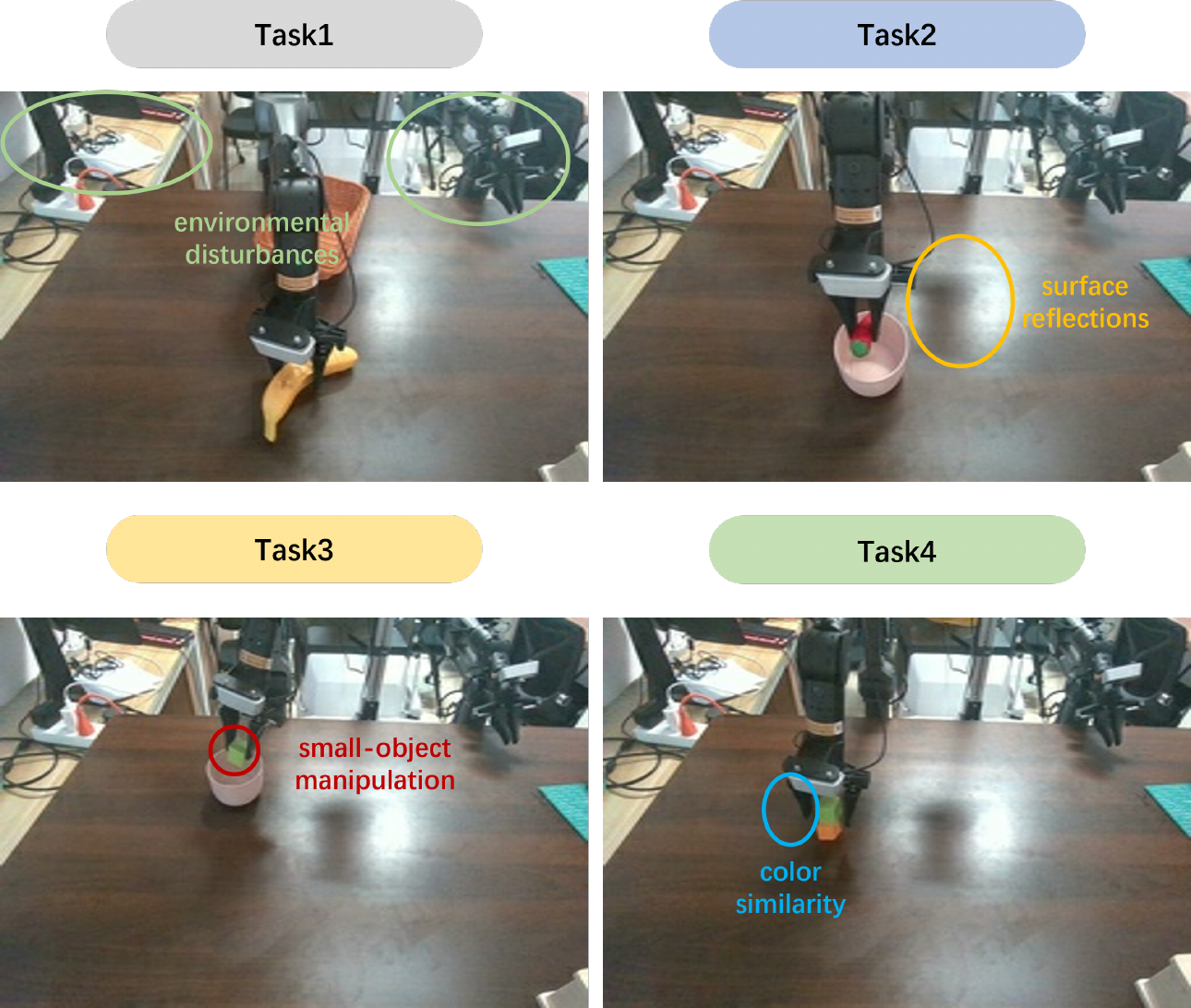}
    \caption{An overview of the main camera view in our real-world task. The environment presents significant challenges for policy learning, including complex lighting conditions, low visual contrast between the gripper and tabletop, and various environmental disturbances that can obscure critical geometric details. }
    \label{fig:real_task}
    \Description{Main camera view of the real-world robotic manipulation setup. A robotic arm with a gripper is positioned above a tabletop workspace. The scene contains objects placed on the table surface. Lighting conditions are uneven, creating shadows and reflections. The gripper has low visual contrast with the tabletop, making it difficult to distinguish boundaries. Background elements and environmental variations introduce additional visual noise that can obscure geometric details.}
\end{figure}

\subsection{Real-Robot Experiments}
\begin{table}
\centering
\caption{Real-World Evaluation on the Piper Arm}
\label{tab:Real Results}
\begin{tabular}{l|cccc|c}
\toprule
 & \textbf{task1}  & \textbf{task2}  & \textbf{task3} & \textbf{task4} & \textbf{Avg} \\
\midrule
ACT \cite{Zhao2023LearningFB} & 20.0 & 0.0 & 0.0 & 0.0 & 5.0\\
Open $\pi_0$ \cite{Ren2024OpenPiZero} &50.0 &40.0 &40.0 &0.0 &32.5 \\
\textbf{QDepth-VLA (ours)}  & \textbf{70.0} &40.0 &\textbf{50.0} &\textbf{10.0} &\textbf{42.5}\\ 
\bottomrule
\end{tabular}
\end{table}
\subsubsection{Environment Setup}
In the real-world experiments, we employ a 6-DoF Piper robotic arm, with a RealSense D455 camera positioned directly in front of the arm. The training hyperparameters are kept consistent with those used in the simulation experiments, except that the action chunk size is set to 16 to achieve faster execution speed. We select four tasks for evaluation: 
\begin{itemize}
\item{\textbf{Task1:} pick the banana into the yellow basket} 
\item{\textbf{Task2:} put the chili into the bowl} 
\item{\textbf{Task3:} put the green block into the bowl}
\item{\textbf{Task4:} stack the green block on top of the yellow block}
\end{itemize}

\begin{table*}[t]
\centering
\caption{Ablation study of QDepth-VLA on Simpler tasks. 
The table reports success rates (\%) with module ablations. 
A checkmark (\checkmark) indicates the module is present, while a cross (\xmark) indicates it is removed, changed or set to zero manually.}
\label{tab:ablation}
\renewcommand{\arraystretch}{1.2}
\setlength{\tabcolsep}{3.5pt} %
\begin{tabular}{l|cccc|cccc|c}
\toprule
\multirow{2}{*}{\textbf{Model}} & 
\multirow{2}{*}{\textbf{Depth Loss}} & 
\multirow{2}{*}{\textbf{Depth Expert}} & 
\multirow{2}{*}{\textbf{Latent Pred}} & 
\multirow{2}{*}{\textbf{Hybrid Attn}} & 
\multicolumn{4}{c|}{\textbf{Tasks}} & 
\multirow{2}{*}{\textbf{Avg}} \\
\cmidrule(lr){6-9}
& & & & & Carrot & Eggplant & Spoon & Block & \\
\midrule
QDepth-VLA         & \checkmark & \checkmark & \checkmark & \checkmark  & 57.5& 95.0 & 82.0 & 39.6 & 68.5 \\
w/o Depth Loss            & \xmark     & \checkmark & \checkmark & \checkmark  & 47.9 & 82.5 &89.2 & 42.9 & 65.6     \\
w/o Depth Expert          & \checkmark & \xmark     & \checkmark & \checkmark  & 61.3 &89.6 &73.7 & 15.8 & 60.0 \\
w/o Latent Prediction       & \checkmark & \checkmark & \xmark     & \checkmark & 54.6 & 80.4 & 82.0 & 41.3 & 64.6 \\
w/o Hybrid Attn \cite{Zhang2025DreamVLAAV}     & \checkmark & \checkmark & \checkmark & \xmark    & 41.7 & 89.6 & 78.8 & 42.0 & 63.0 \\
\bottomrule
\end{tabular}
\end{table*}

For each task, we collect 50 trajectories and fine-tune the model separately on the corresponding dataset. During testing, each task is evaluated over 10 trials.As shown in Figure ~\ref{fig:real_task}, all experiments are conducted on a dark-colored wooden desk, where the surface color is visually similar to the robot gripper. This setup increases the difficulty of the tasks by introducing additional perceptual ambiguity.

\subsubsection{Main Results}
We evaluate QDepth-VLA on a series of pick-and-place tasks with varying difficulty to assess its spatial perception and localization ability. We also compare our method against representative baselines such as ACT  \cite{Zhao2023LearningFB}. As shown in Table \ref{tab:Real Results}, QDepth-VLA consistently outperforms ACT \cite{Zhao2023LearningFB} across all tasks, while ACT \cite{Zhao2023LearningFB} fails to perform reliably in these complex real-world environments, QDepth-VLA achieves robust success rates. Compared to our baseline Open $\pi_0$  \cite{Ren2024OpenPiZero}, QDepth-VLA achieves a 20.0\% improvement on the simple task of picking a banana, and further achieves gains of 10.0\% on both Task 3 and Task 4, demonstrating stronger generalization to challenging scenarios and tasks.

\subsection{Ablation Study}
To evaluate the contribution of each proposed component, we perform a series of controlled ablation experiments in the Simpler  \cite{Li2024EvaluatingRR} simulation environment.
Four ablated variants are considered: (1) removing the depth supervision signal by setting its loss weight to zero (\textbf{w/o Depth Loss}); (2) removing the dedicated depth prediction branch (\textbf{w/o Depth Expert}); (3) replacing latent depth token prediction with pixel-wise regression (\textbf{w/o Latent Prediction}); and (4) substituting the proposed hybrid attention mask with a standard version that enforces proprioception-to-depth attention (\textbf{w/o Hybrid Attn}).
Quantitative results are reported in Table~\ref{tab:ablation}.

\subsubsection{w/o Depth Loss}  
In this variant, the depth loss weight is set to zero while preserving the full model capacity. This configuration isolates the contribution of the depth supervision signal without altering the overall parameter scale.

As shown in Table~\ref{tab:ablation}, performance decreases from 68.5\% to 65.6\% on average. The degradation is most pronounced in the Carrot (-9.6\%) and Eggplant (-12.5\%) tasks, both requiring coarse spatial grounding.
Conversely, slight improvements are observed in Spoon (+7.2\%) and Block (+3.3\%) tasks.
Notably, in the Block task, the depth difference between successful and unsuccessful stacking outcomes can be small, as partially misaligned or unstable stacks may still exhibit similar depth profiles. As a result, depth prediction errors do not necessarily correlate strongly with task failure, and removing the auxiliary depth objective can occasionally simplify optimization and yield modest performance gains.

Overall, these results suggest that depth supervision provides informative depth cues that are beneficial for spatial reasoning beyond the primary control objective.

\subsubsection{w/o Depth Expert}  
Eliminating the dedicated depth branch results in the largest overall performance degradation (-8.5\%), as presented in Table~\ref{tab:ablation}.
The most significant drop occurs in the Stack Block Task (-23.8\%), where precise 3D alignment is critical.
Substantial declines are also observed in the Eggplant (-5.4\%) and Spoon (-8.3\%) tasks, indicating that fine-grained spatial reasoning relies heavily on an explicit and specialized depth pathway.

\subsubsection{w/o Latent Prediction}  
As shown in Table~\ref{tab:ablation}, replacing latent depth prediction with direct pixel-wise regression lowers average performance to 64.6\% (-3.9\%). The largest impact is on Eggplant (-14.6\%) task, whereas other tasks are only mildly affected. This validates our design choice: quantized latent tokens encourage abstraction of geometric cues, while pixel prediction entangles the model with redundant local detail that is less relevant for manipulation.  

\subsubsection{w/o Hybrid Attention}  
In this variant, the proposed hybrid attention mask is replaced with a DreamVLA-style \cite{Zhang2025DreamVLAAV} configuration, removing dynamic and semantic modalities.
This setting tests whether relative depth maps can enhance proprioceptive state perception and thereby improve action generation quality.
As expected, the performance declines by 5.5\% on average, with the most substantial drop on the Carrot Task (-15.8\%).
This result indicates that enforcing proprioception-to-depth attention introduces noise rather than useful guidance, as relative depth lacks absolute positional encoding necessary for stable control.

\subsubsection{Analysis of the Computational Cost}
QDepth-VLA introduces a modest and well-contained computational overhead in terms of model size, runtime, and data annotation. Compared to Open $\pi_0$ \cite{Ren2024OpenPiZero}, the parameter count increases from 2.606B to 2.924B (+12.2\%), mainly due to the additional depth expert and its projection layers, while the core vision-language-action backbone and policy head remain unchanged. At inference time, we observe only minimal runtime overhead in real-robot experiments, with wall-clock control latency remaining comparable to Open $\pi_0$ \cite{Ren2024OpenPiZero}. Depth supervision is obtained via Video-Depth-Anything-Large \cite{Chen2025VideoDA} as an offline preprocessing step, requiring approximately 4 hours to annotate the Bridge and Fractal datasets. The added computational cost represents a favorable trade-off given the consistent gains in spatial reasoning and task success.

Overall, we find the following answers to the three research questions brought up at the beginning of this section,  
\begin{itemize}
  \item Depth supervision effectively enhances VLA performance, especially on long-horizon and fine-grained pick-and-place tasks. In particular, tasks such as stacking and precise placement benefit significantly, indicating improved spatial reasoning.  
  \item Compared to pixel-level regression, quantized depth supervision proves more effective, as it reduces redundancy and focuses learning on salient geometric structures, leading to more stable training and stronger downstream performance.  
  \item The proposed hybrid attention mask consistently contributes to performance gains, particularly in placement tasks, by selectively routing depth cues into the policy network and improving cross-modal feature alignment.  
\end{itemize}

\section{CONCLUSION} 
In this paper, we introduced QDepth-VLA, a new vision-language-action model that incorporates depth supervision and hybrid attention to enhance spatial perception and long-horizon reasoning. Through extensive experiments in both simulation (Simpler and LIBERO) and real-world manipulation tasks, we demonstrate that depth supervision significantly improves manipulation  performance. In summary, our work demonstrates that predicting quantized depth tokens at the current timestep is an effective way to enhance policy learning. Extending this approach to predict future depth tokens for improved reasoning and exploring more efficient VAE-based depth representations for enhanced perception present two promising directions for future research.

\begin{acks}
This work was supported by the National Natural Science Foundation of China (NSFC) under Grants 62136008 and 62293545; in part by the Beijing Major Science and Technology Project under Contract no.Z251100008125023.
\end{acks}




\balance
\bibliographystyle{ACM-Reference-Format} 
\bibliography{sample}


\end{document}